\documentclass{article}
\usepackage{ijcai19}

\usepackage{times}
\usepackage{soul}
\usepackage{url}
\usepackage[hidelinks]{hyperref}
\usepackage[utf8]{inputenc}
\usepackage[small]{caption}
\usepackage{graphicx}
\usepackage{amsmath}
\usepackage{booktabs}
\usepackage{algorithm}
\usepackage{algorithmic}
\usepackage{graphicx}
\usepackage{lipsum}
\usepackage{graphicx}
\usepackage{diagbox}
\usepackage{subfigure}
\usepackage{graphicx}

\title{Asynchronous "Events" are Better For Motion Estimation}

\author{
Yuhu Guo$^{1,2}$\and
Han Xiao$^{1,2}$\footnote{Corresponding Author}\and
Yidong Chen$^{1}$\and
Xiaodong Shi$^{1}$
\affiliations
$^1$ Department of Intelligence Science and Technology, \\ Xiamen University, Fujian Province, 361005, PR China,\\
$^2$ School of Information Science and Technology, Xiamen University Malaysia, Malaysia\\
\emails
SWE1709118@xmu.edu.my, bookman@xmu.edu.cn
}
\begin{document}

\maketitle

\begin{abstract}
Event-based camera is a bio-inspired vision sensor that records intensity changes (called “events”) asynchronously in each pixel. As an instance of event-based camera, Dynamic and Active-pixel Vision Sensor (DAVIS) combines a standard camera and an event-based camera. However, traditional models could not deal with the event stream asynchronously. To analyze the event stream asynchronously, most existing approaches accumulate events within a certain time interval and treat the accumulated events as a synchronous frame, which wastes the intensity change information and weakens the advantages of DAVIS. Therefore, in this paper, we present the first neural asynchronous approach to process event stream for event-based camera.
Our method asynchronously extracts dynamic information from events by leveraging previous motion and critical features of gray-scale frames. To our best knowledge, this is the first neural asynchronous method to analyze event stream through a novel deep neural network. Extensive experiments demonstrate that our proposed model achieves remarkable improvements against the state-of-the-art baselines. 
\end{abstract}

\section{Introduction}
Event-based cameras are novel bio-inspired vision sensors such as Dynamic and Active-pixel Vision Sensor (DAVIS) \cite{Brandli:200837}. Compared to traditional cameras that capture gray-scale frames at a fixed time interval, event-based cameras record the asynchronous event when a single pixel intensity changes. Thus, the output of event-based cameras is the event stream rather than the gray-scale images. For the most conventional event-based camera, event is represented as $(x, y, t, p)$ where $x, y$ indicate the position of the pixel, $t$ is the time stamp and $p \in {-1,+1}$ indicates the polarity that brightness increase or decrease.
Thus, event-based cameras gain the low-latency that $>1\mathrm{kHz}$ compared to traditional cameras (e.g. with $30\mathrm{fps}$). Besides, event-based cameras also show superiority in terms of dynamic range, low bandwidth (e.g. $>120\mathrm{dB}$), high temporal resolution, low storage capacity, low processing time and power consumption \cite{DBLP:journals/corr/abs-1711-01458}. To jointly exploit the advantages of both event-based camera and traditional camera, the Dynamic and Active-pixel Vision Sensor (DAVIS)\cite{Brandli:200837} has been presented in recent years.

\begin{figure}
  \centering

  \begin{tabular}{@{}c@{}}
    \includegraphics[width=\linewidth,height=94pt]{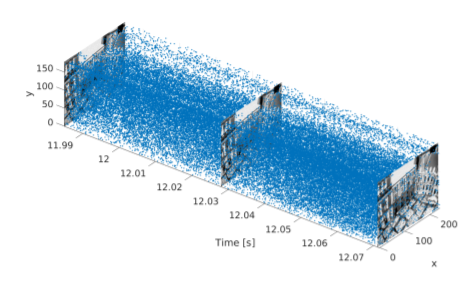} \\[\abovecaptionskip]
   Visualization of the events sequence from DAVIS
  \end{tabular}
  
  \caption{
  DAVIS camera is an enhanced sensor, which combines a standard camera and an event-based camera. Therefore, it provides gray-scale frames (i.e. images) and the events occurring between the gray-scale frames. Specifically, the image slices indicate the gray-scale images that are recorded by the standard camera in a fix rate. The blue dots indicate the events that are recorded by the event-based camera. The events denote the brightness changes for the corresponding pixels in the time gap between two images, which provides the advantages of DAVIS such as low-latency, high-dynamic range, high temporal resolution etc. With the advantages of this novel sensor, we can promote computer vision task effectively.
  }\label{fig:myfig}
\end{figure}

\begin{figure*}[ht]
  \centering 
  \subfigure[APS]{ 
    \includegraphics[width=1.5in]{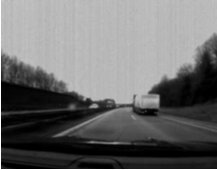} 
  } 
  \subfigure[Event frame(50ms)]{ 
    \includegraphics[width=1.5in]{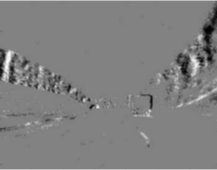} 
  } 
  \subfigure[$h ^ { + } - h ^ { - }$]{ 
    \includegraphics[width=1.5in]{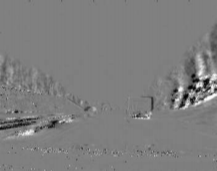} 
  } 
  \subfigure[ our mask]{ 
    \includegraphics[width=1.5in]{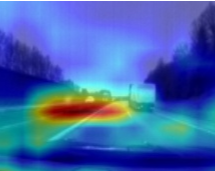} 
  }
  
  \subfigure[APS]{ 
    \includegraphics[width=1.5in]{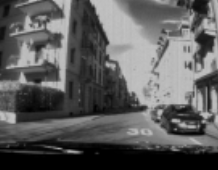} 
  }
  \subfigure[Event frame(50ms)]{ 
    \includegraphics[width=1.5in]{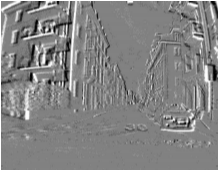}
  }
  \subfigure[$h ^ { + } - h ^ { - }$]{ 
    \includegraphics[width=1.5in]{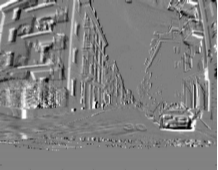} 
  }
  \subfigure[ our mask]{ 
    \includegraphics[width=1.5in]{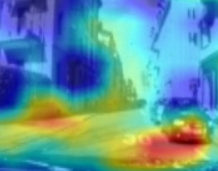}
  }
  
\caption{Visualization of Different Frames. There are two independent cases that the top and bottom lines. {\textbf {APS}} (i,e, (a),(e)) is the original gray-scale images provided by the standard camera of DAVIS. {\textbf{ Event frame (50ms)}} (i.e. (b),(f)) is the accumulated events within the time interval of $50ms$. {\textbf{{$h ^ { + } - h ^ { - }$}} } (i.e. (c), (g)) is developed by Maqueda to collect more information from the accumulated events. {our mask } (i.e. (d), (h)) is the attention mask of our model in the form of heat map.} 
\label{fig:subfig} 
\end{figure*}

\begin{figure*}[ht]
\centering
\includegraphics[scale=0.6]{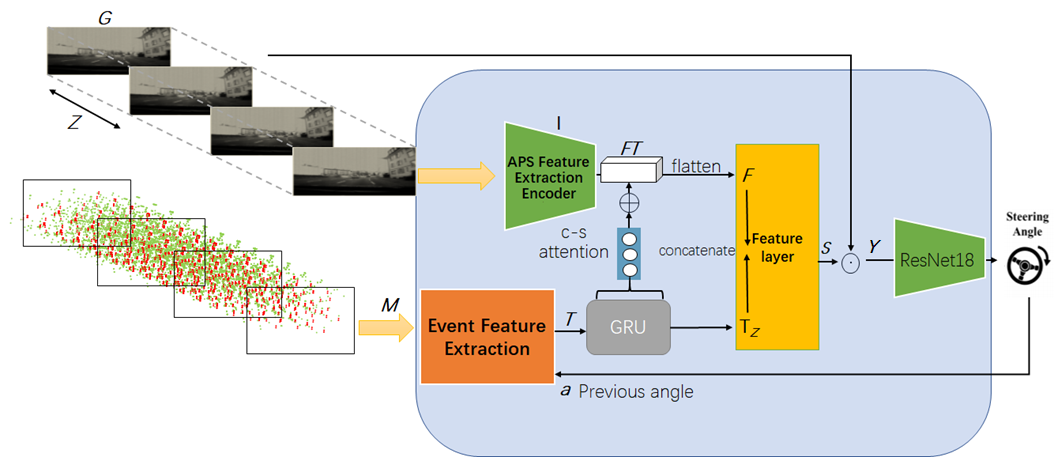}
\caption{Proposed Neural Architecture. First, gray-scale images $G$ are encoded into image-specific feature tensor $I$, while we construct the event matrix $M$ under the same time stamp and leverage event feature extraction module to compress the event matrix $M$ into timestamp-specific vector $T$. Second, GRU  processes the timestamp-specific vectors $T$ in the sequence of time stamp to achieve the hidden representation $h$ for each time stamp. Third, with the input of the hidden representations $h$ in the fix-rate time interval, we leverage the channel-wise and spatial-wise attention mechanism  to address the image-specific tensor $I$ into image-specific feature tensor $FT$. Fourth, we concatenate the flatten image-specific feature tensor $F$ and time interval aligned timestamp-specific vector $T_Z$ as the input of feature layer, while the feature layer applies MLP to produce a mask $S$ to cover the original gray-scale image $G$ as masked image $Y$. Last, we employ ResNet  to map the masked image $Y$ into the steering angle as the target in self-driving task.}
\end{figure*}

DAVIS contains an event-based camera and a standard cameras. Thus, the data stream of DAVIS consists asynchronous event stream and synchronous gray-scale low-rate frames (i.e. images) in fixed time interval. Given the advantages introduced in the first paragraph, event-based cameras outperform standard cameras in multiple tasks, such as motion estimation, feature extraction and object tracking.
Regarding the asynchronous events of event-based cameras, existing models could not fully develop the potentials of event-based camera, because most of them adopt a simple solution, which accumulates events as a new frame in a certain time interval  \cite{E.Mueggler-ICRA-2015}. Recently, inspired by \cite{7508476}, there are several researchers who attempt to split the synchronous accumulated frames into two parts \cite{Maqueda_2018_CVPR,Zhu-RSS-18}, which corresponds to brightness and darkness events respectively.  Demonstrated in Fig.\ref{fig:subfig}, (b) and (c) are the different methods to accumulate the events into fix-rate frames \cite{Maqueda_2018_CVPR}. From the two sub-figures, we observe many redundant information such as trees and clouds, according to the corresponding gray-scale image (a). The redundant information disturbs the performance promotion, which is the critical issue of current models.

Therefore, in this paper, we present a novel neural architecture to asynchronously analyze the event stream. Specifically, first, gray-scale images are encoded into image-specific feature tensor, while we construct the event matrix under the same time stamp and leverage event feature extraction module to compress the event matrix into timestamp-specific vector. Second, GRU \cite{cho-al-emnlp14} processes the timestamp-specific vectors in the sequence of time stamp to achieve the hidden representation for each time stamp. Third, with the input of the hidden representations in the fix-rate time interval, we leverage the channel-wise and spatial-wise attention mechanism \cite{chen2016sca} to address the image-specific tensor into image-specific feature tensor. Fourth, we concatenate the flatten image-specific feature tensor and time interval aligned timestamp-specific vector as the input of feature layer, while the feature layer applies MLP to produce a mask to cover the original gray-scale image as masked image. Last, we employ ResNet \cite{He2016DeepRL} to map the masked image into the steering angle as the target in self-driving task. \textit{\textbf{It is worth to note that our model deals with events for every timestamp rather than accumulating the events in a fixed time interval, which makes our model an asynchronous framework.}}

Extensive experiments demonstrate our method outperforms the state-of-the-art baselines in motion estimation task on the public benchmark dataset of Event Camera self-driving dataset \cite{DBLP:journals/corr/abs-1711-01458}, which justifies the effectiveness of our asynchronous neural architecture.

To summarize, our contributions are two-fold:
\begin{itemize}
\item We propose a novel asynchronous approach for the event stream to capture the most critical dynamic information from the event-based camera DAVIS. To our best knowledge, this is the first time in published literature to address this issue of asynchronous event analysis in a neural architecture.
\item We leverage the attention mechanism that jointly analyzes asynchronous events and gray-scale images (i.e. APS stream) collected by the DAVIS sensor, which achieves substantial improvements against the state-of-the-art baselines in the task of motion estimation. 
\end{itemize}


\section{Related Work}
Recently, event-based cameras have gained extraordinary improvements against standard cameras in many fields. However, the main challenge for event-based cameras is how to leverage the event sequence. Traditional models do not provide the toolbox to handle the event sequence precisely, because they apply the simple solution that to accumulate events in a certain time interval $\delta t$, to make the analysis process similar to synchronous image frames, \cite{E.Mueggler-ICRA-2015,Maqueda_2018_CVPR}. To collect sufficient events during the interval time, most approaches would apply a large time interval (e.g. $50ms$). Inspired by \cite{7508476}, several researchers attempt to split the synchronous frames into two parts that positive events (brightness increase) and negative events (brightness decrease) respectively. 

\cite{Zhu-RSS-18} presents a 4-channel neural architecture to address optical flow estimation issues, the first two channels of which encode the number of positive and negative events and the last two channels of which encode the timestamp of the most recent positive and negative events at that pixel. 

Current state-of-the-art model \cite{Maqueda_2018_CVPR} splits the synchronous frames into separate histograms for positive and negative events as two channels, which are processed into feature vectors and then feed the feature vector into ResNet to achieve the target of self-driving that steering angle.

 Moreover, synchronous methods increase the latency of events cameras, which is against the low-latency property of event-based cameras. Therefore, according to the DAVIS characteristics, \cite{Gehrig_2018_ECCV} presents an asynchronous approach that combines the events and gray-scale images provided by the DAVIS sensor to track features from the geometric model, which leverages each occurring event. Moreover, this asynchronous model beats the state-of-the-art synchronous models in feature tracking task. And there is previous work that also focuses on tracking features by event-based camera \cite{7758089,7989517}. And recently extensions of popular image-based key point detectors have been developed for event-based cameras\cite{Mueggler17BMVC,7759610}

\cite{7605233} jointly analyzes the gray-scale images and synchronous events by a simple convolutional network. This model is also proved successful on the robot control scenario.
Notably, all of the above mentioned neural networks are trained on synchronous event frames in a certain time interval. 

Furthermore, \cite{A.Amir-2017-CVPR} develops a real-time gesture recognition system with a novel chip of the name TrueNorth. And the ability of event-based cameras to provide rich data for solving pattern recognition problems has been initially shown in \cite{6497055,7010933,8050403}

Regarding our novelty, this paper presents the first deep learning driven framework to analyze the events stream asynchronously. Moreover, compared to synchronous redundant event training frames, we leverage each event to extract dynamic information, which is ignored by the synchronous approaches. Furthermore, we apply the channel-wise and spatial-wise attention mechanism \cite{chen2016sca} for our model and verify the effectiveness of attention methods. 

\section{Method}

\begin{table}[ht]
   \arraystretch{1.3}
    \centering
    \caption{Symbol Table}
    \begin{tabular}{c|c}
\hline Symbol & Meaning \\
\hline
\hline  
$C$& predefined threshold for event-based cameras\\
\hline 
$G$& gray-scale image\\
\hline 
$M$& construct event matrix\\
\hline 
$Z$& the fixed time interval of gray-scale \\
\hline 
$T$& hidden representation of event matrix\\
\hline 
$FT$& image-specific feature tensor after C-S attention\\
\hline
$F$& feature vector after flatten the $FT$ tensor\\
\hline 
$A$ & parameter matrix of linear layer \\
\hline 
$L$ & the linear layer in our Event Feature Extraction \\
\hline 
$S$ & mask image after the MLP layer \\
\hline 
$Y$ & the image after $S$ mask cover gray-scale frame$G$ \\
\hline 

$a$ & the vector of previous angle degree \\
\hline 
\end{tabular}
    \label{tab:symbol}
\end{table}

\subsection{Representation of Event Cameras and Events Stream}
Compared to standard cameras, event-based cameras track intensity changes in each pixel and record the changes when the log intensity changes are larger than the predefined threshold $C$:
\begin{equation}\label{equ:Pixelrecord}
    \log (I_{t+1})-\log (I_t) > C
\end{equation}
where $I(t)$ is the intensity in the timestamp $t$ on the image plane and $C$ is the predefined threshold.

Each event consists four elements namely $x-, y-$ pixel location, time stamp and polarity information:
\begin{equation}\label{equ:events}
   e_k=(x_k,y_k,t_k, p_k)
\end{equation}
where $x_k, y_k$ mean the $k$-th x- and y-position of the event, $t_k$ is $k$-th time stamp for $k$-th event $e_k$, and $p_k \in {\{+1},{-1}\}$ indicates the $k$-th polarity for each event that the brightness change (increase or decrease). Due to the asynchronous properties of the events, it is not easy to extract the important dynamic events that are inputs to the APS feature extraction encoder.

\subsection{Architecture}

\begin{figure}[ht]
    \centering
    \includegraphics[scale=0.35]{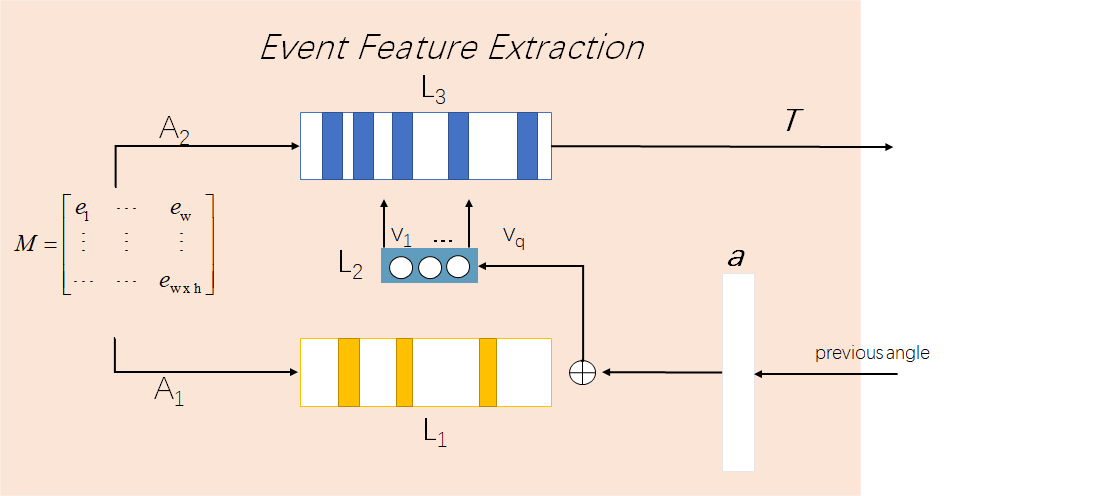}
    \caption{The inputs of this module are the event matrix $M$ and the latest $q$ angles $a = (a_1, .... a_{q})$ where $q$ indicates the number of the latest angles. The output of this module is timestamp-specific vector $T$. 
First, we project the event matrix $M$ with linear layer. Second, we process the output of linear layer with latest angle vector $a$. Third, we project matrix $M$ with another linear layer. Forth, we generate the attention vector from the output of the second step. Last, we achieve the output of this module by column-wise multiply the attention vector and the output of third step.}
    \label{fig4}
\end{figure}

Overall, our neural architectures are composed of five steps. 

First, gray-scale images $G$ with the size of $w \times h$ are encoded into image-specific feature tensor $I \in \mathcal{R}^{w \times h \times c}$ by \textbf{APS feature extraction encoder} where $c$ is the number of channels. Then we construct the \textbf{event matrix} $M \in \mathcal{R}^{w \times h}$ under the same time stamp and leverage \textbf{event feature extraction module} to compress the event matrix into timestamp-specific vector $T \in \mathcal{R}^{q}$ where $q$ is a hyper-parameter. Notably, we will discuss the construction of event matrix and event feature extraction module in next subsections.

Second, GRU processes the $k$-th timestamp-specific vector $T_k$ in the sequence of time stamp to achieve the hidden representation $h_k \in \mathcal{R}^{w}$ for $k$-th time stamp. 

Third, with the input of the hidden representation $\{T_1, ... T_K, ... \}$ in the fix-rate time interval $Z$, we leverage the \textbf{channel-wise and spatial-wise attention mechanism} \cite{chen2016sca} to address the image-specific tensor $I$ into image-specific feature tensor $FT \in \mathcal{R}^{w \times h \times c}$. We flatten the image-specific feature tensor $FT$ into image-specific feature vector $F \in \mathcal{R}^{whc}$.

Then, we concatenate the image-specific vector $F$ and $Z$-th timestamp-specific vector $T_Z$ as the input of feature layer, while the feature layer applies $MLP$ to produce a mask image $S \in \mathcal{R}^{w \times h}$ to cover the original gray-scale image $G$ as masked image $Y \in \mathcal{R}^{w \times h}$. Last, we employ ResNet \cite{He2016DeepRL} to map the masked image $Y$ into the steering angle $D \in \mathcal{R}$ as the target in self-driving task. 

\subsubsection{APS Feature Extraction Encoder}
The input of this module is original gray-scale image $I$, and the output is the image-specific feature tensor $I \in \mathcal{R}^{w \times h \times c}$. The functionality is to extract the hidden features from images. Regarding the detailed structure of this module, please refer to \cite{journals/corr/BojarskiTDFFGJM16}.

\subsubsection{Event Matrix Construction}
The input of event matrix construction module is the event set under the same time stamp, while the output is the polarity matrix $M \in \mathcal{R}^{w \times h}$ with the same size of original gray-scale image $I$. Specifically, if there exist the event $e=(x,y,t,p)$, the entry $(x, y)$ of event matrix $M$ is $p \in \{-1, +1\}$. The other entries that are not recorded in the events are filled with $0$.

\subsubsection{Event Feature Extraction Module}
The inputs of this module are the event matrix $M$ and the latest $q$ angle composed vector $a = (a_1, .... a_{q})$ where $q$ indicates the number of the latest steering angles (the final output of the model). The output of this module is timestamp-specific vector $T$. The functionality is to encode the event in an asynchronous manner. The process is illustrated in Fig.\ref{fig4}.

First, we project the event matrix $M$ with linear layer as 
\begin{equation}
    L_1 = \mathbf{MA_1}
\end{equation}
where $A_1$ is the parameter matrix of this linear layer with the size of $h \times q$. Second, we multiply the output of linear layer with latest angle vector $a$:
\begin{equation}
    L_2 = \mathbf{L_1}a
\end{equation}
Third, we project matrix $M$ with another linear layer:
\begin{equation}
    L_3 = \mathbf{MA_2}
\end{equation}
where $A_2$ is the parameter matrix of this linear layer with the size of $h \times q$.
Forth, we generate the attention vector from $L_2$:
\begin{equation}
    v = \operatorname {softmax} (\mathbf{L_2})
\end{equation}
where $v$ is the attention vector with the size of $q \times 1$.
Last, we achieve the output of this module (i.e. timestamp-specific vector $T$) by applying column-specific multiplication between $v$ and $L3$:
\begin{equation}
    T = L_3 \otimes v
\end{equation}
where $\otimes$ means the column-specific multiplication operation.

\section{Experiments}
\subsection{Performance Metrics}
To follow the previous literature, we employ the root-mean-squared error (RMSE) to metric the performance \cite{Maqueda_2018_CVPR}:
\begin{equation}\label{equ:RMSE}
\mathrm { RMSE } \doteq \sqrt { \frac { 1 } { N } \sum _ { j = 1 } ^ { N } \left( \hat { \alpha } _ { j } - \alpha _ { j } \right) ^ { 2 } }
\end{equation}
where RMSE measures the average magnitude of the prediction error, showing how close the observed values $ {\alpha } $ are to those predicted value $ \hat {\alpha }$  by the model.
We also apply the explained variance (EVA) to evaluate our model stability as \cite{Maqueda_2018_CVPR}
\begin{equation}\label{equ:EVA}
\mathrm { EVA } \doteq 1 - \frac { \operatorname { Var } ( \hat { \alpha } - \alpha ) } { \operatorname { Var } ( \alpha ) }
\end{equation}
where $\operatorname { Var } ( \hat { \alpha } - \alpha )$ is given by the variance of the residuals between observed values $\alpha$ and predicted value $\hat { \alpha }$. EVA measures the proportion of variation in the predicted values with respect to those of the observed values. 
If predicted value $ \hat {\alpha }$ fits the observed value $ {\alpha } $ approximately, the total variance will be greater than the residual variance, which leading to EVA greater than $1$. On the other hand, if the performance is unsatisfactory, the total variance will be equal or less than the residual variance, resulting in EVA equal or less than $0$,respectively.
\begin{equation}\label{equ:averageRmse}
\mathrm { Improvement} \doteq \frac {  ( \hat { \beta } - \beta ) } {  \beta  }\%
\end{equation}
where $\beta$ is the original RMSE and $\hat { \beta }$ is the RMSE, which we want to compared with original one. We will employ the average Improvement in our Results \& Analysis subsection in 4 scenarios to compared with different methods directly.

\subsection{Datasets}
We apply the public benchmark dataset \cite{DBLP:journals/corr/abs-1711-01458} for our experiments. The dataset contains over 12 hours driving recordings, which are collected by vehicles under real and challenging scenarios. Given the fact that the data are collected by DAVIS, the dataset consists of asynchronous events and gray-scale images (APS), along with vehicle speed, GPS position, driver steering and throttle. \cite{Maqueda_2018_CVPR} segmented the recordings into four subsets, namely day, day sun, evening and night, according to the weather and scenarios. The duration of recordings ranges from minutes to hours. And most of steering angles are slight deviations of $\pm 10$ degrees. Also the speed is uniformly distributed over the range $0-160 km/h$. Notably, subsets differ in not only the weather conditions and illumination, but also in the travelled route. For example, there is number of turns on city and town, but hardly exists on the high way scenarios. 
\begin{table*}[t]
    \centering
    \renewcommand\arraystretch{1.3}
    \begin{tabular}{|l|c|c|c|c|} \hline
     \diagbox{\bfseries Scenarios}{\bfseries RMSE(EVA)}{\bfseries Methods} & {\bfseries Maqueda(APS)}  & {\bfseries Maqueda(ResNet18)}   &{\bfseries Maqueda(ResNet50)}  &   {\bfseries Asynchronous(Ours) } \\ 
      \hline
      \hline
      day  &  4.57 (0.047)  & 2.99 (0.551) & 2.33 (0.728)  & \bfseries 2.17 (0.812) \\ \hline
      day sun  &20.07 (0.125)  &  10.87 (0.742) & 9.47 (0.805)  & \bfseries 8.05 (0.875)\\ \hline
      evening  &7.23 (0.172)  & 5.45 (0.518)  & 5.01 (0.602)  & \bfseries 4.67 (0.734)\\ \hline
      night  & 6.96 (0.181) &4.51 (0.654)   & \bfseries 3.82 (0.753)  & 3.94 (0.711) \\ \hline
    \end{tabular}
    \caption{
    Comparison with synchronous learning approaches by using gray-scale(APS) frames as well as event frames, for the each scenarios. The APS baseline is based on ResNet 18 network.(best results per row are highlighted in bold)
    } \label{tab:TimeEfficency}
\end{table*}

\subsection{Implementation}

Similarly as previous study, we also split the data to four parts according to the scenarios: day, day sun, evening and night \cite{DBLP:journals/corr/abs-1711-01458}. \textbf{Compared with the state-of-the-art baseline, we apply the same dataset segment, the same pre-processing and the same tricks, which makes our experiments a fair comparison.}



We have tested many experimental settings and achieve the optimal setting as: $w=260$, $h=346$, $q=256$, $Z=10$ in $10fps$ dataset or $Z=50$ in $50fps$. Besides, our model is trained by ADAM \cite{Kingma_2014}, with the hyper-parameter settings:$\beta _ { 1 } = 0.9$, $\beta _ { 2 } = 0.999$ and $\epsilon = 10 ^ { - 8 }$, and our initial learning is $0.0001$.

We will release our codes and the documents for the codes upon acceptance.

\subsection{Results \& Analysis}
Regarding the experimental results in Table 2, we could study two critical questions:

\begin{enumerate}
    \item How to prove that our asynchronous approach is better than traditional synchronous approaches for the event stream information?
    \item Why our asynchronous approach is better at extracting dynamic information?
\end{enumerate}

For a fair comparison, we deploy the same ResNet18 or ResNet50 networks as the feature encoders in our model as two experimental settings.  We can observe that if we only apply gray-scale image as the input, both RMSE and EVA would be bad. Obviously, the average RMSE is different among different sets, because the RMSE is dependent on the absolute value of the steering ground truth.

On the other hand, our model outperforms the only gray-scale image input, approximately improved over $33.34$\% in average. The effectiveness of our model stems from to filter the redundant information such as the clouds or the backgrounds. Specifically, in our method, we employ the same gray-scale images, but we design an attention-mask to cover the gray-scale images by element-wise multiplication operation to filter out the redundant information. Thus, Our asynchronous approach also beats all the baselines in the settings of ResNet18 over about $19.5$\% in average. The results demonstrate that the performance improvements stem from the attention masks which are generated in an asynchronous framework. In this way, we answer the first critical question that our asynchronous approach is better than traditional methods that are synchronous.

Notably, in our asynchronous approach, the setting of ResNet50 outperforms the settings of ResNet18 in the sub-dataset of day, day sun, and evening scenarios over approximately $6.25$\% (average in RMSE), which accords to our common sense. Therefore, it is reasonable to conjecture that to leverage the more effective CNN network, such as ResNet101 or inception, we could get a much better performance. 

\begin{figure}[htbp]
\label{fig:Case}
\centering
\subfigure[day.]{
\begin{minipage}[t]{0.5\linewidth}
\centering
\includegraphics[width=1in]{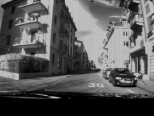}
\end{minipage}%
}%
\subfigure[day sun.]{
\begin{minipage}[t]{0.5\linewidth}
\centering
\includegraphics[width=1in]{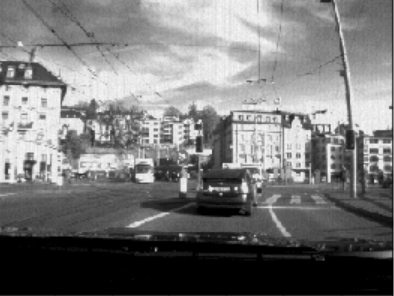}
\end{minipage}%
}%

\subfigure[evening.]{
\begin{minipage}[t]{0.5\linewidth}
\centering
\includegraphics[width=1in]{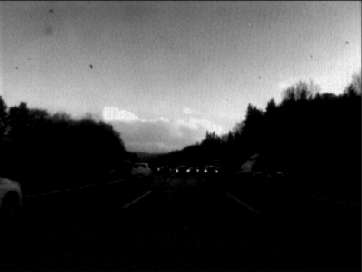}
\end{minipage}
}\subfigure[dark night.]{
\begin{minipage}[t]{0.5\linewidth}
\centering
\includegraphics[width=1in]{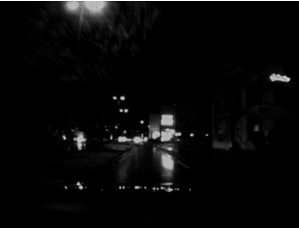}
\end{minipage}
}
\centering
\caption{Sample input gray-scale images extracted from the dataset for four scenarios.}
\end{figure}

Note in the dataset of night, the baseline model of ResNet50 leads the performance a bit. The main reasons behind why our ResNet18-based model can not beat ResNet50-based baseline in night scenarios are listed:
\begin{itemize}
    \item Our final network is simple, which ResNet18 is more simple than ResNet50. But noting that we beat the ResNet18-based baseline which is the comparative model, our method indeed outperforms the baselines. To consider that our model beats the ResNet50-based baseline in other three datasets, our model outperforms the baseline extensively.
    \item \cite{DBLP:journals/corr/abs-1711-01458} remarks that most of gray-scale images at night are too dark. The figure 5 (d) shows that it is hardly to capture clear images at night dataset relative to other three sub figures, because of low brightness intensity. Thus, the darkness of night would also affect the event-based camera for recording the events. Although our method is good at capturing the important events at night, after our attention-mask covers the gray-scale images, it still can not gain sufficient information, compared to the other scenarios.
    \item Besides, most of night situation driving at free way or city, is under a high speed. Due to the flaw of standard camera at high velocities, the gray-scale images would get blurred. Regarding our model, we leverage the generated attention mask to process the blurred gray-scale images, which leads to unsatisfactory performance
\end{itemize}

Regarding the second question, we generate the visualization of different frames in each method based on the codes of \cite{Kim2017InterpretableLF}. As we mentioned, DAVIS camera combines a standard camera and an event-based camera. Thus, the differences of two cameras could be observed from Figure 2 (a) and (b). 
The  traditional gray-scale images can capture all scenes. But, it cannot attend to dynamic points at high velocities. In the comparison, event-based cameras can directly capture the dynamic points from the scenes. In this way, it ignores a lot of non-critical still information, which is the reason why the event-based camera can handle motion estimation task under a high speed object.

However, one individual time stamp is not sufficient to collect information, so previous studies present the method that accumulates the events in a certain time interval $h ^ { + } - h ^ { - }$ \cite{Maqueda_2018_CVPR} to avoid information loss. Although their methods are efficient, as you can observe from Figure 2 (b) and (c) that it still has the useless events (noise). To address this issue, we present our asynchronous attention-based approach to select the most representative points. 
To compare with all the baselines, the visualization of our attention masks over the APS frame is demonstrated in Figure 2 (d) and (h), where red color indicates dense part while blue color indicates sparse part. From the visual results, it is concluded that our mask rearranges the dynamic points captured by our network. Besides, most of selected points focus on the specific object and edge which are the most critical information of self-driving, such as cars, houses and route lines. Specifically, compared to the synchronous approaches, our method ignores tree, sky and clouds, which are noise for self-driving task. In conclusion, our asynchronous approach is better at extracting dynamic information because it can avoid the redundant information and noise to address the critical points.

\section{Conclusion}
In this paper, we show an attention-based asynchronous approach for self-driving task. Our method leverages the event stream asynchronously and extracts dynamic points for the sense without redundant information.Moreover, we leverage the attention mechanism that jointly analyzes asynchronous events and gray-scale images to achieve substantial improvements over the state-of-the-art baselines in the task of motion estimation. Experiments prove that our asynchronous approach can beat synchronous approaches extensively.

\newpage

\bibliographystyle{named}
\bibliography{ijcai19}

\end{document}